\documentclass[conference]{IEEEtran}
\ifCLASSINFOpdf
\else
\fi
\usepackage{times}
\usepackage{soul}
\usepackage{url}
\usepackage[hidelinks]{hyperref}
\usepackage[utf8]{inputenc}
\usepackage{graphicx}
\usepackage{amsmath}
\usepackage{amsthm}
\usepackage{booktabs}
\usepackage{algorithm}
\usepackage{algorithmic}
\usepackage[switch]{lineno}

\usepackage{xspace}
\usepackage{comment}
\usepackage{multirow}
\usepackage[normalem]{ulem}
\usepackage{enumitem}
\usepackage{booktabs}
\usepackage{amssymb}
\usepackage{colortbl} %
\usepackage{xcolor}   %
\usepackage{arydshln} %

\definecolor{red3}{HTML}{f03c1d}
\definecolor{blue2}{HTML}{4266ad}
\definecolor{blue3}{HTML}{2e4da3}
\definecolor{yellow3}{HTML}{fca00a}
\definecolor{hotpink}{HTML}{ff69b4}

\newcommand{\method}[0]{\textsc{Crayon}\xspace}
\newcommand{\methodattention}[0]{\textsc{Crayon-Attention}\xspace}
\newcommand{\methodpruning}[0]{\textsc{Crayon-Pruning}\xspace}
\newcommand{\methodall}[0]{\textsc{Crayon-Pruning+Attention}\xspace}
 
\newcommand{\yesnof}[0]{yes-no annotations\xspace}

\definecolor{green4}{HTML}{2f670b}
\definecolor{green3}{HTML}{0f9d58}
\definecolor{green2}{HTML}{70a74c}
\definecolor{green1}{HTML}{7bcfa9}
\definecolor{green0}{HTML}{b3ec8b}
\newcommand{\best}[1]{\textbf{#1}}

\renewcommand{\sectionautorefname}{\S\kern-1pt}
\renewcommand{\subsectionautorefname}{\S\kern-1pt}
\renewcommand{\subsubsectionautorefname}{\S\kern-1pt}
\renewcommand{\sectionautorefname}{Sec.}
\renewcommand{\subsectionautorefname}{Sec.}
\renewcommand{\subsubsectionautorefname}{Sec.}

\begin{document}

\title{%
Effective Guidance for Model Attention with Simple Yes-no Annotations}

\author{
\IEEEauthorblockN{Seongmin Lee}
\IEEEauthorblockA{
Georgia Institute of Technology\\
Email: seongmin@gatech.edu}
\and
\IEEEauthorblockN{Ali Payani}
\IEEEauthorblockA{Cisco Systems Inc.\\
Email: apayani@cisco.com}
\and
\IEEEauthorblockN{Duen Horng (Polo) Chau}
\IEEEauthorblockA{
Georgia Institute of Technology\\
Email: polo@gatech.edu}
}

\maketitle

\begin{abstract}
Modern deep learning models often 
make predictions by focusing on irrelevant areas, leading to biased performance and limited generalization.
Existing methods aimed at rectifying model attention
require explicit labels for  irrelevant areas
or complex pixel-wise ground truth attention maps.
We present \method (\underline{C}orrecting \underline{R}easoning with \underline{A}nnotations of \underline{Y}es \underline{O}r \underline{N}o), 
 offering 
effective, scalable, and practical solutions to 
rectify model attention
using simple \yesnof.
\method empowers classical and modern model interpretation techniques  
to identify and guide model reasoning:
\methodattention directs classic interpretations based on saliency maps
to focus on relevant image regions, while 
\methodpruning removes irrelevant neurons identified by modern concept-based methods to mitigate their influence.
Through extensive experiments with both quantitative and human evaluation, we showcase CRAYON’s effectiveness, scalability, and practicality in refining model attention.
CRAYON achieves state-of-the-art performance, outperforming 12 methods across 3 benchmark datasets,
surpassing approaches that require more complex annotations.
\end{abstract}

\IEEEpeerreviewmaketitle

\section{Introduction}
\label{sec:intro}
Deep learning models have achieved remarkable performance, even surpassing humans in tasks such as image classification~\cite{he2015delving}.
However, recent advancements in deep learning interpretation have discovered that
these models often make predictions focusing on irrelevant areas~\cite{singla2021salient,xiao2020noise}. %
For example, a model trained to classify waterbirds and landbirds often bases its predictions on \textit{backgrounds}, not bird bodies,
as waterbirds often appear near bodies of water like lakes, while landbirds are commonly found near land features like forests~\cite{sagawa2019distributionally}.
Such model attention to less relevant areas results in
reduced trustworthiness~\cite{kaur2022trustworthy},
poor generalization~\cite{oakden2020hidden,rodolfa2021empirical}, and
biased performance~\cite{wang2020towards,srinivasan2021biases,seo2022information,wang2022fairness}.
Thus, it is crucial to improve these models to attend to pertinent areas~\cite{oakden2020hidden,rodolfa2021empirical}.
Existing methods to mitigate irrelevant attentions suffer from major drawbacks.
Some techniques attempt to decorrelate class labels from irrelevant areas by explicitly annotating each image's irrelevant regions.
For example, in a bird dataset, images may be annotated with either  water or land backgrounds,
which are then used to 
balance the training dataset~\cite{li2019repair,nam2020learning,yao2022improving} or
reduce correlations between irrelevant background areas and class labels~\cite{kim2019learning,liu2022avoiding,nam2020learning}.
However, in practice, due to the wide variety of %
irrelevant areas, categorizing them can be challenging~\cite{xiao2020noise,sarridis2023flac}.
Other methods directly guide model attention using ground truth attention maps to inform the model where it should focus or avoid focusing, in order to align the model's saliency maps with these ground truth maps~\cite{ross2017right,gao2022aligning,gao2022res}.
However, these methods necessitate accurate pixel-wise maps for every training data point, which can be excessively time-consuming and labor-intensive. %
Also, 
discrepancies between real-valued model-generated maps and binary ground truth maps can degrade performance~\cite{gao2022res}.

\begin{figure}[t]
    \centering
    \includegraphics[width=0.98\columnwidth]{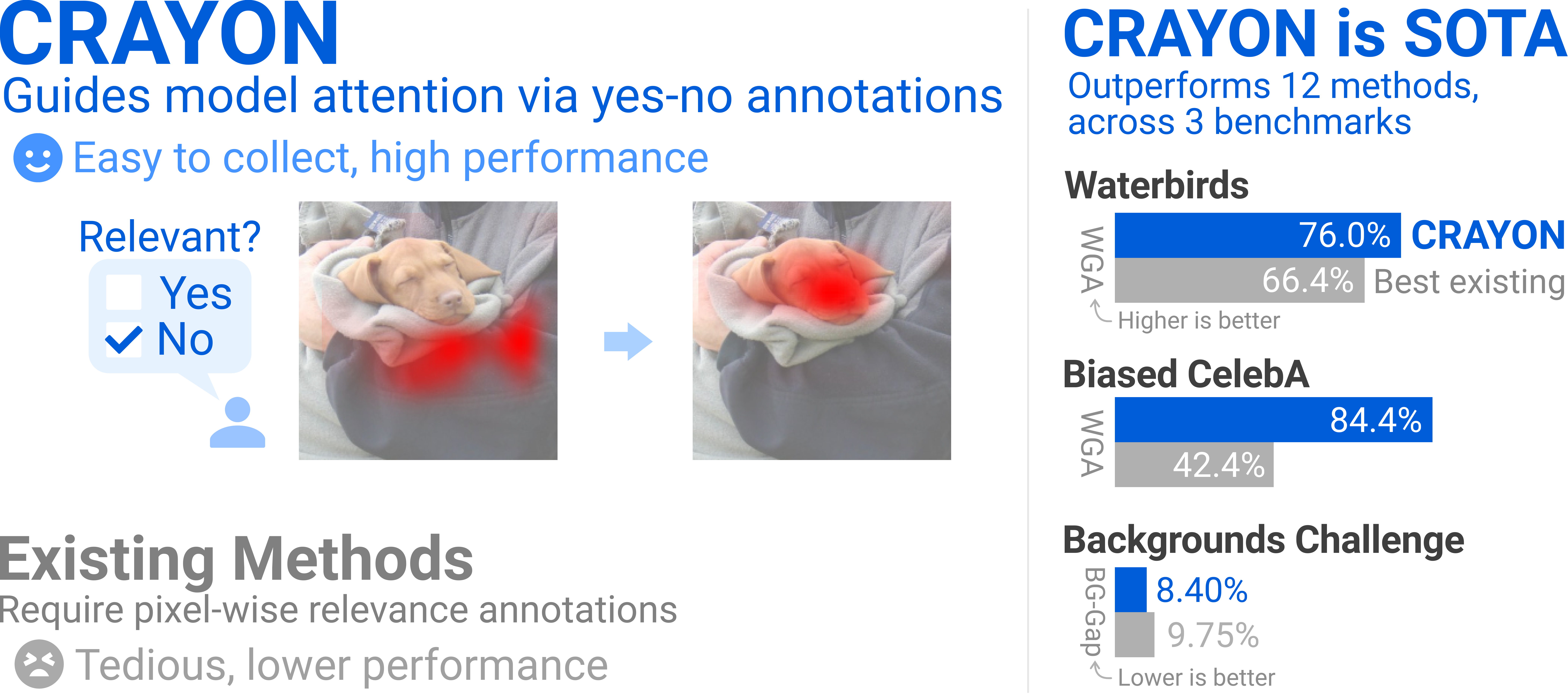}
    \vspace{-3pt}
    \caption{
    \textbf{Left:} \method uses purposefully simple \yesnof{} to guide a model to attend to relevant image areas 
    (e.g., redirect from background to foreground), 
    overcoming limitations of existing methods requiring complex annotations such as pixel-wise maps.
    \textbf{Right:}
    \method achieves state-of-the-art performance, surpassing 12 other methods across 3 benchmark datasets. 
    }
    \label{fig:crownjewel}
    \vspace{-10pt}
\end{figure}

To address the above research gaps, we present \textbf{\method} (\underline{C}orrecting \underline{R}easoning with \underline{A}nnotations of \underline{Y}es \underline{O}r \underline{N}o),
which makes the following contributions:
\begin{itemize}[leftmargin=15pt] %
    \item 
    \textbf{Yes-No Annotations: An Effective Strategy to Guide Model Attention.}
    Our research shows that such purposefully simple annotations 
    is a highly effective and practical solution for correcting model attention, 
    overcoming critical limitations of existing methods that require laborious annotations,
    while also delivering superior performance. %
    \method empowers both classic saliency map-based and modern concept-based interpretations 
    to not only identify but also rectify model attention. (\autoref{sec:methods})
    \item\textbf{\method achieves state-of-the-art performance in
    correcting model attention,
    outperforming 12 existing methods across 3 benchmark datasets.} 
    Through extensive quantitative experiments with large-scale human annotations involving almost 6,000 participants, we showcase \method{}'s effectiveness, scalability, and practicality in refining model attention.
    Remarkably, \method surpasses all methods that require more complex annotations and
    achieves near-peak performance with annotations for just 10\% of the training data.
    To promote reproducible research and transparency, we open-source our code at  \url{https://github.com/poloclub/crayon}.
    (\autoref{sec:evaluation})
\end{itemize}

\section{Related Work}
\label{sec:related}
A lot of efforts have been dedicated to rectifying the attention of deep learning models.
Some researchers have attributed irrelevant attention to spurious correlations in training data~\cite{sagawa2020investigation}
and alleviate such issues by reweighting or subsampling training data~\cite{li2019repair,nam2020learning,liu2021just,kirichenko2022last,xue2023eliminating}.
However, challenges arise when a training dataset lacks spurious-free data. 
In response, some researchers opt to create balanced training datasets
by collecting or generating additional instances~\cite{lee2021learning,kim2021biaswap,chiu2022better,yao2022improving,han2022umix}. %
Yet, these approaches can be costly or impractical in real-world scenarios~\cite{schwartz2022limitations}.
Various loss terms~\cite{kim2019learning,singla2021salient,zhang2021causaladv}
and pruning approaches~\cite{nagpal2020attribute} have been introduced to counter the impact of spurious correlations.
However, all these methods require annotations for the attributes that are spuriously correlated with the class labels,
while identifying %
the correlated attributes is challenging in practice~\cite{xiao2020noise}.
Several methods addressing such limitations have shown some efficacy~\cite{liu2021just,asgari2022masktune,liu2022avoiding,zhang2022correct}.

To achieve higher performance while overcoming all the aforementioned limitations,
researchers have incorporated humans in refining vision models~\cite{kulesza2015principles,teso2019explanatory,schramowski2020making}.
Various approaches have extended model interpretation techniques beyond mere identification, aiming to align model attributions with human intuition~\cite{saha2023saliency,schneider2023reflective}. %
Many researchers have focused mainly on saliency maps and collected human annotations for model attention.
The RRR loss~\cite{ross2017right} was proposed to 
redirect MLP models away from regions annotated by humans as irrelevant,
later extending its applicability to deeper models~\cite{gao2022aligning,gao2022res}.
CDEP~\cite{rieger2020interpretations} and SPIRE~\cite{plumb2021finding} aim to reduce the impact of irrelevant pixels by leveraging contextual decomposition and masking specific objects in images, respectively.
Stammer et al.~\cite{stammer2021right} refine models at both pixel and concept levels by disentangling concepts within an image.
However, all these methods require humans to supply ground truth attention maps for each image, which can be prohibitively costly to obtain.
To address this challenge, some progress has been made by using eye-gaze tracking apparatus~\cite{ma2023eye} or
introducing simpler alternatives such as scribble maps~\cite{shen2021human} and 
bounding boxes~\cite{rao2023studying},
which yet do not resolve the inherent limitations of human-provided attention maps~\cite{gao2022res}.
We further purposefully simplify human feedback to \yesnof on model interpretation results, overcoming existing methods' limitations while delivering superior performance.

\section{Methods}
\label{sec:methods}
\subsection{Overview}
\label{sec:methods-overview}
\method fine-tunes a trained model %
to base its predictions on relevant data regions
by harnessing simple \yesnof{}, which pertain to
the relevance of the rationale behind the model's predictions,
as revealed through model interpretations.
In this section, we describe
(1) how \yesnof for classic interpretations based on saliency maps guide the model's attention to the relevant regions --- we call this  \methodattention (\autoref{sec:crayon-attention})
and
(2) how we extend our idea to modern concept-based interpretations to prune the neurons activated by irrelevant visual concepts --- we call this \methodpruning (\autoref{sec:crayon-pruning}).

\subsection{CRAYON-Attention: Guide Saliency Maps}
\label{sec:crayon-attention}
Generating saliency maps stands as one of the most commonly employed and deeply explored model interpretation techniques~\cite{selvaraju2017grad}. Therefore, we chose to leverage it as a familiar mechanism to collect human annotations in our novel and simple way.
For a given model and its training data $\mathbf{x}_1,\dots,\mathbf{x}_N$,
the saliency map $M_{\mathbf{x}_n}$ highlights the regions within the image $\mathbf{x}_n$ that the model focuses on for its prediction.
Once saliency maps are generated for all
$N$ training data points, 
we proceed to gather \yesnof regarding the relevance of each map to the prediction task.
We denote the set of indices corresponding to training data with relevant and irrelevant maps as ${R}$ and ${I}$, respectively.

To refine the model using the \yesnof, we introduce a loss function based on the energy loss~\cite{wang2020score}.
For the data point $\mathbf{x}_n$ whose saliency map $M_{\mathbf{x}_n}$ highlights the relevant regions (i.e., $n\in R$),
the model should generate similar saliency maps following the refinement.
Hence, we formulate the loss function $\mathcal{L}_{rel,n}$ as follows:
\begin{equation}
\label{eq:rel_loss}
    \mathcal{L}_{rel,n}=
    \sum_{h=1}^H{
    \sum_{w=1}^W{
        [M'_{\mathbf{x}_n}]_{hw}
        (1-[M_{\mathbf{x}_n}]_{hw})
    }
    }
\end{equation}
where $H$ and $W$ represent the height and width of the saliency maps, respectively,
and
$M'_{\mathbf{x}_n}$ is the saliency map for the model being trained and the data point $\mathbf{x}_n$.
We clarify that $M_{\mathbf{x}_n}$ is the saliency map for the original model before refining, and 
$M'_{\mathbf{x}_n}$ is for the model being fine-tuned.
For better stability of the loss function,
we normalize both $M_{\mathbf{x}_n}$ and $M'_{\mathbf{x}_n}$, scaling their values between 0 and 1 by dividing each map by its maximum value.

For the data point $\mathbf{x}_{n}$ with irrelevant saliency map (i.e., $n\in I$),
the model should attend to the regions that are not highlighted in the map $M_{\mathbf{x}_n}$.
In this regard, we construct the loss function $\mathcal{L}_{irrel,n}$ as follows:
\begin{equation}
\label{eq:irrel_loss}
    \mathcal{L}_{irrel,n}=
    \sum_{h=1}^H{
        \sum_{w=1}^W{
            [M'_{\mathbf{x}_n}]_{hw}
            [M_{\mathbf{x}_n}]_{hw}
        }
    }
\end{equation}

While guiding the model to attend to the right regions, we need to preserve the accuracy of the model's predictions.
Therefore, we incorporate the prediction loss $\mathcal{L}_{pred,n}$ for the data point $\mathbf{x}_n$:
\begin{equation}
\label{eq:predloss} 
    \mathcal{L}_{pred,n}=
    \sum_{k=1}^K{
    -y_{nk}\log{\hat{y}_{nk}}
    }
\end{equation}
where 
$y_{nk}$ is 1 if the label of the data $\mathbf{x}_n$ is $k$ and 0 otherwise
and
$\hat{y}_{nk}$ is the probability of the data $\mathbf{x}_n$ being labeled as $k$ computed by the model being trained.

Summing up the loss functions, we obtain the loss $\mathcal{L}_{att}$ that guides a model with \yesnof on saliency maps,
\begin{equation}
\label{eq:attloss}
    \mathcal{L}_{att}=
    \sum_{n=1}^{N}{\mathcal{L}_{pred,n}}+
    {\frac{\alpha}{\lvert R\rvert}}\sum_{n\in {R}}{\mathcal{L}_{rel,n}}+
    {\frac{\beta}{\lvert I\rvert}}\sum_{n\in {I}}\mathcal{L}_{irrel,n}
\end{equation}
where $\alpha$ and $\beta$ are the hyperparameters that control the weights of the loss terms,
and $\lvert R\rvert$ and $\rvert I\rvert$ are the number of relevant and irrelevant training data points respectively.

\subsection{CRAYON-Pruning: Prune Irrelevant Neurons}
\label{sec:crayon-pruning}
\begin{figure}
  \centering
  \includegraphics[width=\columnwidth]{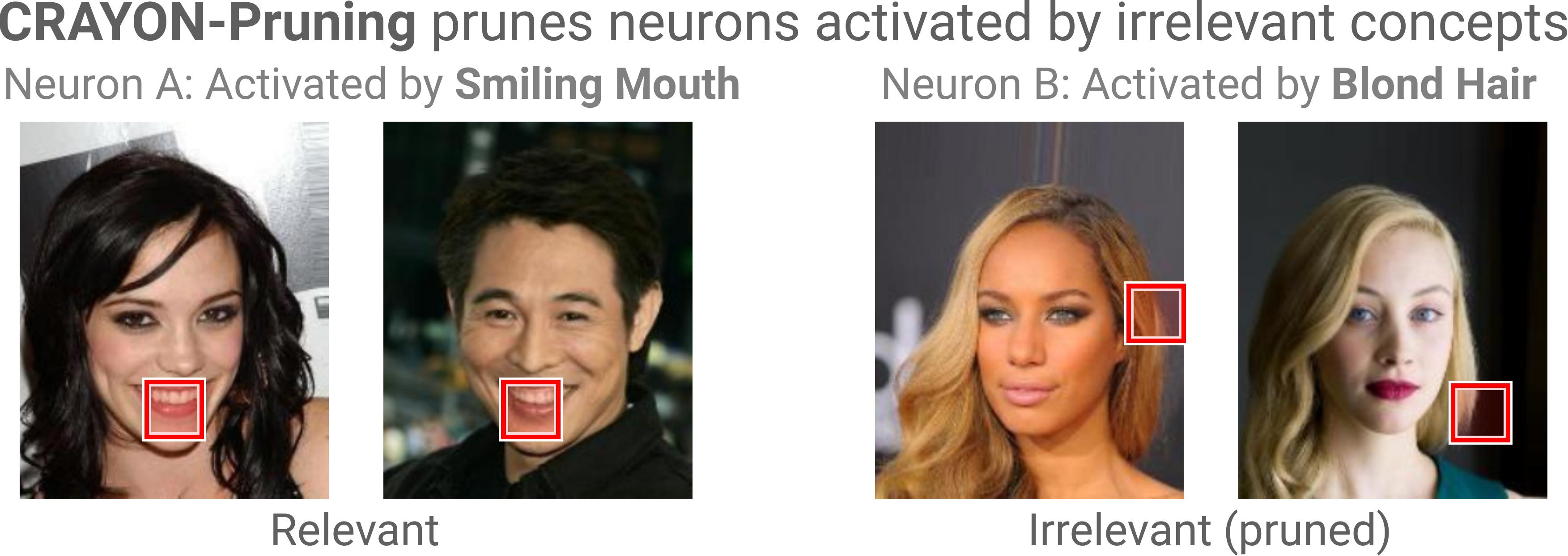}
  \caption{
  \methodpruning prunes the neurons activated by irrelevant concepts in the penultimate layer and fine-tunes the last layer.
  For example, in a smile classifier,
  \textbf{Left:} a neuron is activated by \textit{smiling mouth}, which is relevant to smiling, while
  \textbf{Right:} another neuron %
  is activated by irrelevant \textit{blond hair} and is pruned. %
  }
  \label{fig:prune}
\end{figure}

Neurons, also referred to as \textit{channels}, 
in the penultimate layer of CNN models are %
activated by specific high-level visual concepts in the input data~\cite{bengio2013representation}.
Based on this finding,
a model interpretation method
that summarizes the concepts responsible for a neuron's activation as a collection of image patches has  recently been proposed~\cite{hohman2020summit}.
These patches are identified by selecting the images that most highly activate the neuron and cropping out the corresponding region. 
For example,
a neuron in the penultimate layer of a smile classifier 
has patches corresponding to the \textit{mouth} concept, indicating that the neuron's activation is attributed to the presence of a mouth (\autoref{fig:prune}, left),
while another neuron has patches for \textit{hair} (\autoref{fig:prune}, right).

\methodpruning identifies the neurons in the penultimate layer that are activated by irrelevant concepts by presenting the image patches of each neuron and collecting \yesnof on their relevance.
For instance, for the smile classifier in the previous example,
the neuron activated by the \textit{mouth} is relevant,
while the neuron activated by the \textit{hair} is irrelevant.
We prune the irrelevant neurons and fine-tune the last fully-connected layer of the model to remove the effect of the irrelevant concept on the model's prediction.
For this fine-tuning process, we use the prediction loss in \autoref{eq:predloss}.

\subsection{Time Complexity}
\label{sec:complexity}
Both \methodattention and \methodpruning share the same general time complexity of $\mathcal{O}(NEM)$, 
where $N$ is the number of training data points, 
$E$ is the number of training epochs, and
$M$ corresponds to time for loss computation.
For \methodattention, the loss computation is based on Grad-CAM (as described in \autoref{sec:crayon-attention}), whereas 
for \methodpruning, it uses the conventional cross-entropy loss.
This analysis indicates that the runtimes of both \methodattention and \methodpruning scale linearly with the number of training data points, as we will empirically demonstrate in our experiments in \autoref{sec:scalability}.

\section{Evaluation with Human Annotations}
\label{sec:evaluation}
To demonstrate  \method's effectiveness, scalability, and practicality in rectifying model attention, 
we collect yes-no human annotations for three benchmark datasets from 5,893 participants on Amazon Mechanical Turk (MTurk).

\subsection{Benchmark Datasets and Model Training}
\label{sec:dataset}
\noindent
\textbf{Waterbirds}~\cite{sagawa2019distributionally}
consists of bird images, where
waterbirds and landbirds are more commonly seen in water (e.g., lake) and land (e.g., forest) backgrounds, respectively.
Its training set consists of:
\begin{itemize} [itemsep=1mm, leftmargin=15pt] %
    \item 1,057 \textit{waterbirds} on \textit{water} backgrounds,
    \item 3,498 \textit{landbirds} on \textit{land} backgrounds,
    \item 56 \textit{waterbirds} on \textit{land} backgrounds, and
    \item 184 \textit{landbirds} on \textit{water} backgrounds.
\end{itemize}
Bird classifiers trained on this data tend to classify birds based on the backgrounds rather than their bird features.
The test set consists of 1,284 waterbirds and 4,510 landbirds; half of the waterbird images and half of the landbird images have water backgrounds, while the other half have land backgrounds.

\smallskip 
\noindent
\textbf{Biased CelebA}~\cite{krishnakumar2021udis}
demonstrates a correlation
where most training images of individuals with \textit{black hair} exhibit \textit{smiling} expressions, while those with \textit{blond hair} are mostly \textit{not smiling}.
There are a total of 20,200 training instances:
\begin{itemize} [itemsep=1mm, leftmargin=15pt] %
\item 10,000 with \textit{black hair} and \textit{smiling} attributes,
\item 10,000 with \textit{blond hair} and \textit{not smiling} attributes,
\item 100 with \textit{black hair} and \textit{not smiling} attributes, and
\item 100 with \textit{blond hair} and \textit{smiling} attributes.
\end{itemize}
This correlation %
leads the smile classifier trained on this dataset to make incorrect associations between smile predictions and hair color.
The test set contains a total of 8,000 data instances, with 2,000 instances per group.

\smallskip 
\noindent
\textbf{Backgrounds Challenge}~\cite{xiao2020noise}
addresses the problem that classifiers trained on the ImageNet~\cite{deng2009imagenet} dataset often inappropriately base their predictions on image backgrounds, rather than foreground objects.
Aiming to correct models to base their predictions on the foreground objects, the challenge introduces the ImageNet-9 (IN-9) dataset, a subset of ImageNet with nine coarse-grained classes (e.g., dog, bird, vehicle).
The IN-9 dataset's training and test set consist of 5,045 and 450 images for each class, respectively (45,405 and 4,050 in total).

\medskip
For each dataset, we train a ResNet50~\cite{he2016deep} classifier, pretrained with ImageNet~\cite{deng2009imagenet},
with the cross-entropy loss (\autoref{eq:predloss}) for 10 epochs,
and refer to it as the \textit{Original} model throughout this paper, indicating no refinement has been applied and thus it may attend to irrelevant areas.
We train the \textit{Original} model for the Waterbirds dataset
with Adam optimizer~\cite{kingma2014adam} with a learning rate of 0.001, a weight decay of 0.0001, and a batch size of 256 for 50 epochs. Images are resized to 256$\times$256, center-cropped to 224$\times$224, and then normalized.
For the Biased CelebA, we use Adam optimizer with a learning rate of 0.001, a weight decay of 0.0001, and a batch size of 128 for 10 epochs.  Images are resized to 274$\times$224.
For the Backgrounds Challenge, we use the model provided in the challenge repository as the Original model. Images are resized to 256$\times$256, center cropped to 224$\times$224, and then normalized.

\subsection{Annotating Visualized Interpretations}
\label{sec:annotation}
\begin{figure} %
  \centering
  \includegraphics[width=0.95\columnwidth]{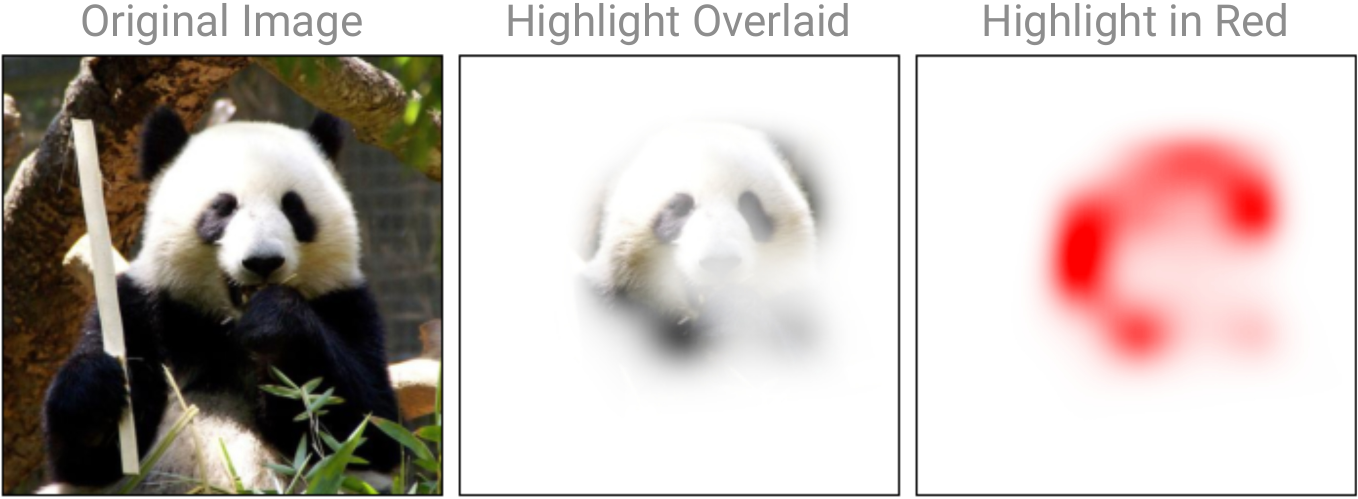}
  \caption{Example visualization shown to participants for attention annotations.} %
  \label{fig:vis-examples}
\end{figure}

We collect \yesnof for the relevance of saliency maps and neuron concepts on Amazon MTurk. %
To collect human annotations for \methodattention, we present participants with a visualization for each training image, as the example shown in \autoref{fig:vis-examples}, consisting of:
\begin{itemize} [topsep=4pt, itemsep=1mm, parsep=1pt, leftmargin=15pt] %
  \item \textit{Original} training image
  \item \textit{Highlight Overlaid} depicting regions receiving higher Grad-CAM attention with higher opacity
  \item \textit{Highlight in Red} coloring model-attended areas in red
\end{itemize}
This visualization design was informed by our pilot study, where we learned that relying solely on \textit{Highlight in Red}, which has conventionally been used~\cite{adebayo2018sanity}, could obscure image contents, making it hard for participants to assess relevance. 
Similarly, identifying model-attended areas solely with \textit{Highlight Overlaid} is challenging when the attended areas are white in color.
We present all three visualization components for Waterbirds and Backgrounds Challenge,
while only \textit{Highlight Overlaid} is presented for Biased CelebA, as
most areas in these face images are in skin colors, and many participants have found it sufficient to determine relevance.
To maximize quality and efficiency, we present each visualization to two participants~\cite{welinder2010online} 
and ask each participant a single yes-no question:
\begin{itemize} [topsep=4pt, itemsep=1mm, parsep=1pt, leftmargin=15pt] %
  \item Waterbirds: Is the strong highlight mainly on the bird? %
  \item Biased CelebA: Can you determine %
  if the person in the image is smiling? %
  \item Backgrounds Challenge\footnote{
  As the Backgrounds Challenge is a large dataset with 45,045 training images, we collected annotations incrementally and discovered that annotating a small random subset of just 10\% (4,500) was already sufficient to achieve SOTA results, as presented in \autoref{sec:performance}.}: Is the strong highlight mainly on X? (X is the image's class label, e.g., dog, wheeled vehicle, fish)
\end{itemize}

We carefully set our questions to include the word ``mainly'' for the Waterbirds and Backgrounds Challenge datasets so that the participants would pick ``\textit{yes}'' only if they strongly agree with the question, rather than just checking for the existence of highlight.
Images that receive ``\textit{yes}'' responses (e.g., highlight only on the bird for Waterbirds) 
from both participants are annotated as having a \textit{relevant} saliency map.
Conversely, if both respond with ``\textit{no},'' the image is annotated as \textit{irrelevant}.
Images receiving mixed responses, suggesting possible Grad-CAM ambiguity, are not used to guide the model's attention. %
For \methodpruning, 
we generate image patches that represent the visual concepts of each neuron in the model's penultimate layer.
Since a neuron's concepts are often determined by a combination of multiple patches~\cite{hohman2020summit},
we generate three patches for each neuron, following convention~\cite{rashtchian2010collecting}.
These patches are created by identifying the top three data points that yield the highest activation for the neuron, along with the activation's spatial position.
As an image patch could strongly activate multiple neurons, we eliminate such duplicates, resulting in
a set of 2,060 unique patches for the Waterbirds dataset, 4,627 for the Biased CelebA dataset, and 3,042 for the Backgrounds Challenge dataset.
We visualize each patch by overlaying a red rectangle on the corresponding position of the image (\autoref{fig:prune}).
Each patch is shown to an individual participant, and we inquire 
whether the red highlight is on the bird body for the Waterbirds dataset and 
whether the area covered by the patch implies the presence of a smile for the Biased CelebA dataset.
For the Backgrounds Challenge dataset, we ask whether the red highlight is on X, where X is the image's class label.
A neuron is annotated as \textit{relevant} if the majority of its patches receive \textit{yes} responses,
and as \textit{irrelevant} if the majority receive \textit{no} responses. 

\subsection{\method Configurations}
\label{sec:crayonsetup}
We outline the hyperparameter configurations for \method used in our experiments. %
For the Waterbirds dataset, \methodattention fine-tunes the \textit{Original} bird classifier for 10 epochs with a batch size of 128, using the Adam optimizer~\cite{kingma2014adam} with a learning rate of 5e-5 and a weight decay of 1e-4. We set the hyperparameters $\alpha$ and $\beta$ in \autoref{eq:attloss} set to 1e+7 and 2e+5, respectively.
For Biased CelebA, the \textit{Original} smile classifier is fine-tuned for 10 epochs with a batch size of 64, using the Adam optimizer with a learning rate of 1e-5 and a weight decay of 1e-4. We set the hyperparameters $\alpha$ to 5e+7 and $\beta$ to 1e+6.
For Backgrounds Challenge, we fine-tune the classifier for 10 epochs with a batch size of 256 using the SGD optimizer~\cite{amari1993backpropagation} with a learning rate of 5e-6 and a weight decay of 1e-1. The hyperparameters $\alpha$ and $\beta$ set to 5e+3 and 5e+2, respectively.

For \methodpruning, we prune 1,034 irrelevant neurons from the penultimate layer of the \textit{Original} model trained on the Waterbirds dataset, and then fine-tunes the last fully connected layer for 10 epochs with a learning rate of 5e-5.
For the Biased CelebA dataset, we prune 1,871 irrelevant neurons and train the last layer for 50 epochs with a learning rate of 5e-6.
For the Backgrounds Challenge dataset, 407 neurons are pruned, and the last layer is trained for 10 epochs with a learning rate of 1e-6.

We also evaluate \methodall, which prunes irrelevant neurons in the penultimate layer and then fine-tunes the entire model with attention guidance\footnote{
Neuron pruning must be applied before using attention annotations, as  annotated image patches stem from  \textit{Original} model. Fine-tuning the model with \methodattention{} may change the concepts detected by each neuron and new image patches may appear.}.
For Waterbirds, we set $\alpha$ to 1e+7 and $\beta$ to 2e+5,
while for Biased CelebA, we set $\alpha$ and $\beta$ to 5e+7 and 1e+6, respectively.
For Backgrounds Challenge, we set $\alpha$ to 1e+3 and $\beta$ to 5e+1.
For the Waterbirds and Biased CelebA datasets, we use the Adam optimizer with the same learning rate, weight decay, and number of epochs as with \methodattention, while we use the SGD optimizer with a learning rate of 5e-5 and weight decay to 1e-1 for the Backgrounds Challenge dataset.
We select these values base on the performance of training data after testing with a wide range of hyperparameter values (details in \autoref{sec:hyperparam}).

\begin{table*}[t]
\centering
\caption{
\method achieves the \best{best}, state-of-the-art performance in correcting model attention, outperforming all existing methods across 3 benchmark datasets, even surpassing methods that require more complex annotations.
}
\label{tab:perf}
\small
\begin{tabular}{@{}ll@{\hspace{2em}}cc@{\hspace{2em}}cc@{\hspace{2em}}cc@{}}
\toprule
\multirow{2}{*}{Method} & \multirow{2}{*}{Annotation} &  \multicolumn{2}{@{\hspace{-2.25em}}c}{\textbf{Waterbirds}} & \multicolumn{2}{@{\hspace{-2.5em}}c}{\textbf{Biased CelebA}} & \multicolumn{2}{@{\hspace{0.5em}}c}{\textbf{Backgrounds}} \\
\cmidrule(l{-1.5em}){3-8} 
&  & WGA$\uparrow$ & MGA$\uparrow$ & WGA$\uparrow$ & MGA$\uparrow$ & MR$\uparrow$ & BG-Gap$\downarrow$  \\ 
\midrule
Original                            & - & 28.35 & 72.08 & 32.60 & 73.71 & 78.27 & 12.99     \\
[3pt]
\hdashline[1pt/1pt]
\noalign{\vskip 3pt}
\textbf{\methodattention}                    & Yes-No & 72.31\tiny{$\pm$0.89} & 85.23\tiny{$\pm$0.17} & 83.29\tiny{$\pm$2.07} & 89.61\tiny{$\pm$0.10} & 80.85\tiny{$\pm$0.23} & 8.52\tiny{$\pm$0.29} \\
\textbf{\methodpruning}                      & Yes-No & 68.97\tiny{$\pm$0.43} & 83.13\tiny{$\pm$0.08} & 69.51\tiny{$\pm$2.20} & 86.75\tiny{$\pm$0.40} & 78.61\tiny{$\pm$0.15} & 12.18\tiny{$\pm$0.11} \\
\textbf{\methodall}                          & Yes-No & \best{76.04}\tiny{$\pm$3.38} & \best{86.03}\tiny{$\pm$0.45} & \best{84.38}\tiny{$\pm$1.49} & \best{90.13}\tiny{$\pm$0.36} & \best{81.66}\tiny{$\pm$0.14} & \best{8.40}\tiny{$\pm$0.31} \\
[3pt]
\hdashline[1pt/1pt]
\noalign{\vskip 3pt}
JtT~\cite{liu2021just}              & - & 
46.88\tiny{$\pm$1.69} & 78.29\tiny{$\pm$0.92} & 35.25\tiny{$\pm$1.90} & 74.29\tiny{$\pm$0.31} & 77.61\tiny{$\pm$0.20} & 12.99\tiny{$\pm$0.09} \\
MaskTune~\cite{asgari2022masktune}  & - & 
45.67\tiny{$\pm$2.17} & 79.13\tiny{$\pm$0.36} & 37.72\tiny{$\pm$1.37} & 78.85\tiny{$\pm$0.28} & 78.14\tiny{$\pm$0.08} & 12.54\tiny{$\pm$0.07} \\
LfF~\cite{nam2020learning}          & - & 
44.64\tiny{$\pm$0.85} & 77.24\tiny{$\pm$0.27} & 44.35\tiny{$\pm$1.76} & 77.05\tiny{$\pm$0.40} & 78.23\tiny{$\pm$0.05} & 12.41\tiny{$\pm$0.08} \\
SoftCon~\cite{hong2021unbiased}     & - & 
46.10\tiny{$\pm$4.56} & 79.93\tiny{$\pm$1.28} & 42.38\tiny{$\pm$13.99} & 76.17\tiny{$\pm$3.23} & 73.14\tiny{$\pm$2.02} & 11.58\tiny{$\pm$0.75} \\
FLAC~\cite{sarridis2023flac}        & - & 
40.68\tiny{$\pm$10.58} & 80.77\tiny{$\pm$3.59} & 39.31\tiny{$\pm$11.84} & 76.73\tiny{$\pm$3.51} & 79.91\tiny{$\pm$0.36} & 9.75\tiny{$\pm$0.35} \\
LC~\cite{liu2022avoiding}           & - & 
61.65\tiny{$\pm$3.70} & 80.43\tiny{$\pm$0.67} & 64.98\tiny{$\pm$5.09} & 84.80\tiny{$\pm$0.88} & 74.78\tiny{$\pm$1.81} & 13.35\tiny{$\pm$0.55} \\
CnC~\cite{zhang2022correct}         & - & 
46.98\tiny{$\pm$1.12} & 77.80\tiny{$\pm$0.47} & 37.76\tiny{$\pm$0.65} & 75.34\tiny{$\pm$0.19} & 77.87\tiny{$\pm$0.42} & 12.91\tiny{$\pm$0.27} \\
[3pt]
\hdashline[1pt/1pt]
\noalign{\vskip 3pt}
RRR~\cite{ross2017right}            & Map & 
53.96\tiny{$\pm$3.46} & 82.29\tiny{$\pm$0.83} & 42.16\tiny{$\pm$4.22} & 78.64\tiny{$\pm$0.82} & 80.08\tiny{$\pm$0.25} & 10.68\tiny{$\pm$0.22} \\
GradMask~\cite{simpson2019gradmask} & Map & 
58.38\tiny{$\pm$3.83} & 82.78\tiny{$\pm$0.66} &  8.43\tiny{$\pm$6.70} & 65.72\tiny{$\pm$2.45} &  80.11\tiny{$\pm$0.30} & 10.63\tiny{$\pm$0.30} \\
ActDiff~\cite{viviano2019saliency}  & Map & 
64.58\tiny{$\pm$1.66} & 84.54\tiny{$\pm$0.23} & 41.29\tiny{$\pm$4.69} & 79.08\tiny{$\pm$1.20} & 75.84\tiny{$\pm$5.97} & 11.73\tiny{$\pm$3.06} \\
GradIA~\cite{gao2022aligning}       & Yes-No, Map   & 
60.87\tiny{$\pm$2.86} & 83.17\tiny{$\pm$0.51} & 41.80\tiny{$\pm$7.82} & 76.58\tiny{$\pm$2.09} & 79.79\tiny{$\pm$0.19} & 11.26\tiny{$\pm$0.15} \\
Bounding Box~\cite{rao2023studying} & Bounding Box  & 
66.36\tiny{$\pm$2.17} & 85.85\tiny{$\pm$0.46} & 33.34\tiny{$\pm$1.96} & 73.03\tiny{$\pm$0.38} & 80.82\tiny{$\pm$0.13} & 9.81\tiny{$\pm$0.96} \\
[3pt]
\hdashline[1pt/1pt]
\noalign{\vskip 3pt}
ERM         & - & 
22.49\tiny{$\pm$12.00} & 70.23\tiny{$\pm$5.10} & 37.82\tiny{$\pm$9.54} & 71.86\tiny{$\pm$3.46} & 77.86\tiny{$\pm$0.26} & 12.88\tiny{$\pm$0.21} \\
\bottomrule
\end{tabular}
\end{table*}

\subsection{Compared Methods}
\label{sec:human_compared}
To demonstrate the importance of human annotations, we compare \method with some of the most recent, best-performing methods aimed at rectifying model attention without human annotations: %
\begin{itemize} [topsep=4pt, itemsep=1mm, parsep=1pt,  leftmargin=15pt] %
    \item \textit{JtT}~\cite{liu2021just} upweights the loss of  training data points misclassified by the \textit{Original} model.
    \item \textit{MaskTune}~\cite{asgari2022masktune} guides the model to learn diverse features by masking regions highly attended by the \textit{Original} model.
    \item \textit{LfF}~\cite{nam2020learning} introduces an auxiliary model that heavily relies on irrelevant areas to identify the data points that are likely to receive irrelevant attention.
    \item \textit{SoftCon}~\cite{hong2021unbiased} %
    uses the auxiliary model to closely locate data points with the same class labels but significant disparities in irrelevant areas. %
    \item \textit{FLAC}~\cite{sarridis2023flac} 
    maximizes the dissimilarity between the feature distributions of the \textit{Original} model and those of the auxiliary model. %
    \item \textit{LC}~\cite{liu2022avoiding} corrects logits of the \textit{Original} model using the auxiliary model outputs.
    \item \textit{CnC}~\cite{zhang2022correct} 
    replaces the auxiliary model's predictions required by SoftCon with the \textit{Original} model's predictions.
\end{itemize}

\noindent
We also assess five methods that require complex human annotations
by using the \textit{ground truth} maps provided in the dataset.
We note that this comparison poses \method disadvantages
since it uses \textit{simple human-provided} annotations, not \textit{complex ground truth}.
For Biased CelebA, which lacks segmentation maps, we generate maps covering the eyes and mouth, relevant to smiling, as the images in this dataset are aligned with respect to the eyes' locations~\cite{liu2015faceattributes}. 
Bounding boxes are generated to enclose these maps.
It is important to note that we could not collect human annotations for the existing methods because they either require proprietary apparatus~\cite{gao2022res,gao2022aligning} or are not evaluated with human annotators~\cite{ross2017right,rao2023studying}.
The methods\footnote{We tried RES~\cite{gao2022res} on our datasets and determined that it was computationally prohibitive. The algorithm did not finish one iteration even after 3 hours on an NVIDIA A6000 GPU; 2,205 iterations are needed for the Biased CelebA dataset.}  are:

\begin{itemize} [topsep=4pt, itemsep=1mm, parsep=1pt, leftmargin=15pt] %
    \item \textit{RRR}~\cite{ross2017right} collects ground truth maps that annotate irrelevant pixels in the images and guides the model not to attend to these irrelevant pixels.
    \item \textit{GradMask}~\cite{simpson2019gradmask} collects ground truth maps %
    and penalizes attention outside the relevant pixels.
    \item \textit{ActDiff}~\cite{viviano2019saliency} collects ground truth maps to mask the background of each image and closely aligns the masked image's representation with the unmasked one.
    \item \textit{GRADIA}~\cite{gao2022aligning} collects \yesnof to identify images where the original model generates irrelevant saliency maps or incorrect predictions, collects ground truth maps, and aligns the model's attention accordingly.
    \item \textit{Bounding Box}~\cite{rao2023studying} collects bounding boxes that cover the relevant regions of each image and guides the model to keep its attention within these boxes.
\end{itemize}

\noindent
To ensure attention correction is not solely due to extended training, we examine the naive \textit{empirical risk minimization (ERM)}~\cite{sagawa2019distributionally} approach that simply minimizes  classification loss by training the model for more epochs.
Hyperparameter values for all compared methods are determined through a comprehensive systematic search around the values reported in their papers. We use the values that yield the highest performance.
We report each method's training configurations in our  repository\footnote{\url{https://github.com/poloclub/crayon/blob/main/misc/training_config.md}}.

\subsection{Evaluation Metrics}
\label{sec:metrics}
As the Waterbirds and Biased CelebA provide the labels for irrelevant areas,
we employ \textit{worst} and \textit{mean group accuracy} as the evaluation metrics %
in accordance with the practice in literature~\cite{sagawa2019distributionally}.
Specifically,
we first evaluate the model accuracy for each group created by intersecting irrelevant area labels and class labels; %
for example,  Waterbirds consists of four groups:
waterbirds on water backgrounds, waterbirds on land backgrounds, landbirds on water backgrounds, and landbirds on land backgrounds.
We then compute the minimum and mean accuracy values across these groups and 
denote them as \textit{worst group accuracy} (WGA) and \textit{mean group accuracy} (MGA), respectively.

For Backgrounds Challenge, 
we employ the metrics proposed by the challenge itself~\cite{xiao2020noise},
as it is difficult to categorize its image backgrounds to form image groups.
These metrics are based on 
two datasets, \textit{Mixed-Same} and \textit{Mixed-Rand}, which are created by transforming image backgrounds.
The \textit{Mixed-Same} dataset shuffles backgrounds across the images with the same class label, while
the \textit{Mixed-Rand} dataset shuffles the backgrounds across all images %
to decorrelate backgrounds and class labels.
We evaluate:
\begin{itemize} [itemsep=1mm, leftmargin=15pt] %
    \item Accuracy on the Mixed-Rand dataset to examine whether the model makes correct predictions when the class label is decorrelated from background.
    \item Difference between the accuracies on the Mixed-Same and Mixed-Rand datasets to assess the impact of backgrounds on the model predictions; we name it as \textit{BG-Gap}. A smaller BG-Gap implies less reliance on backgrounds.
\end{itemize}

\subsection{Results: \method is Effective and SOTA}
\label{sec:performance}

\method achieves state-of-the-art performance in correcting model attention, outperforming 12 existing methods across three benchmark datasets, including those that require more complex annotations. %
We run each method five times with different random seeds and report the average and standard deviation of the performance values in \autoref{tab:perf}.

For Waterbirds,
compared to the unrefined original model (Row~1),
\methodattention (Row~2) substantially enhances
WGA by 43.96 percentage points (pp), raising it from 28.35\% to 72.31\%;
and
MGA by 13.15pp, %
from 72.08\% to 85.23\%.
\methodpruning (Row~3) also demonstrates improvements, achieving a 40.62pp increase in WGA and an 11.05pp increase in MGA.
\methodall (Row~4), which combines both \textit{attention} and  \textit{pruning} approaches, further elevates both WGA and MGA beyond the capabilities of each individual approach, 
raising WGA to 76.04\% and MGA to 86.03\%.
\method is also effective for Biased CelebA, significantly outperforming all 12 methods. which is expected for the methods that do not leverage human annotations (\autoref{tab:perf}, Row 4-10).
JtT and LfF, which upweight the training data points misclassified by the original model, 
exhibit marginal effects as there are few misclassified training data points due to the severe imbalance in training data. %
Similarly, CnC, which uses misclassified data as positive and negative samples for contrastive learning, achieves only limited improvement.
For the methods that use human annotations, their performances were on par with those without human annotations. We believe this is due to the discrepancies between real-valued model-generated maps and binary ground truth maps~\cite{gao2022res}.
Moreover, we observe that GradMask (Row~13) often fails because its masking significantly alters the face areas other than mouth and eyes,
making it hard for the model to detect the image as a face.
Bounding Box (Row~16) shows limited performance, as rectangular boxes surrounding eyes and mouth inevitably include irrelevant areas like nose and cheeks, where hair frequently appears.

For Backgrounds Challenge,
\method
achieves the highest accuracy on the Mixed-Rand dataset and smallest BG-Gap,
demonstrating its effectiveness in multi-label classification tasks.
It improves the accuracy on the Mixed-Rand dataset from 78.27\% to 81.66\%
and reduces the BG-Gap from 12.99\% to 8.40\%.
While \methodpruning exhibits marginal performance, since a concept irrelevant to one class can be relevant to another in multi-label classification,
combining it with attention guidance
complements such limitation and effectively mitigates the model reliance on image backgrounds.
The underperformance of 
most approaches that do not use any annotations %
demonstrates their limitations in the challenging setting where irrelevant attributes correlated with class labels are unclear~\cite{liu2022avoiding}.
Specifically, the underperformance of JtT and CnC is attributed to the small number (only 2\%) of training data points misclassified by the original model,
while MaskTune's limited performance suggests its potential reliance on large training data, %
as higher performance was reported 
when four times the training data points were used~\cite{asgari2022masktune}.

It is noteworthy that
\methodattention surpasses all methods using more complex annotations, which often contain richer information. %
We attribute \method's superior performance to its ability to overcome the shortcomings of binary ground truth maps and boxes, 
which are represented as 0s and 1s,
while model-generated saliency maps consist of continuous real numbers.
This inconsistency degrades the performance of model attention guidance~\cite{gao2022res}.
Additionally, experiments with the Biased CelebA dataset reveal that the shape constraint of bounding boxes significantly degrades performance.
\method addresses this challenge by using the saliency maps of the original model instead of binary ground truth.

\begin{figure}[t]
    \centering
    \includegraphics[width=\columnwidth]{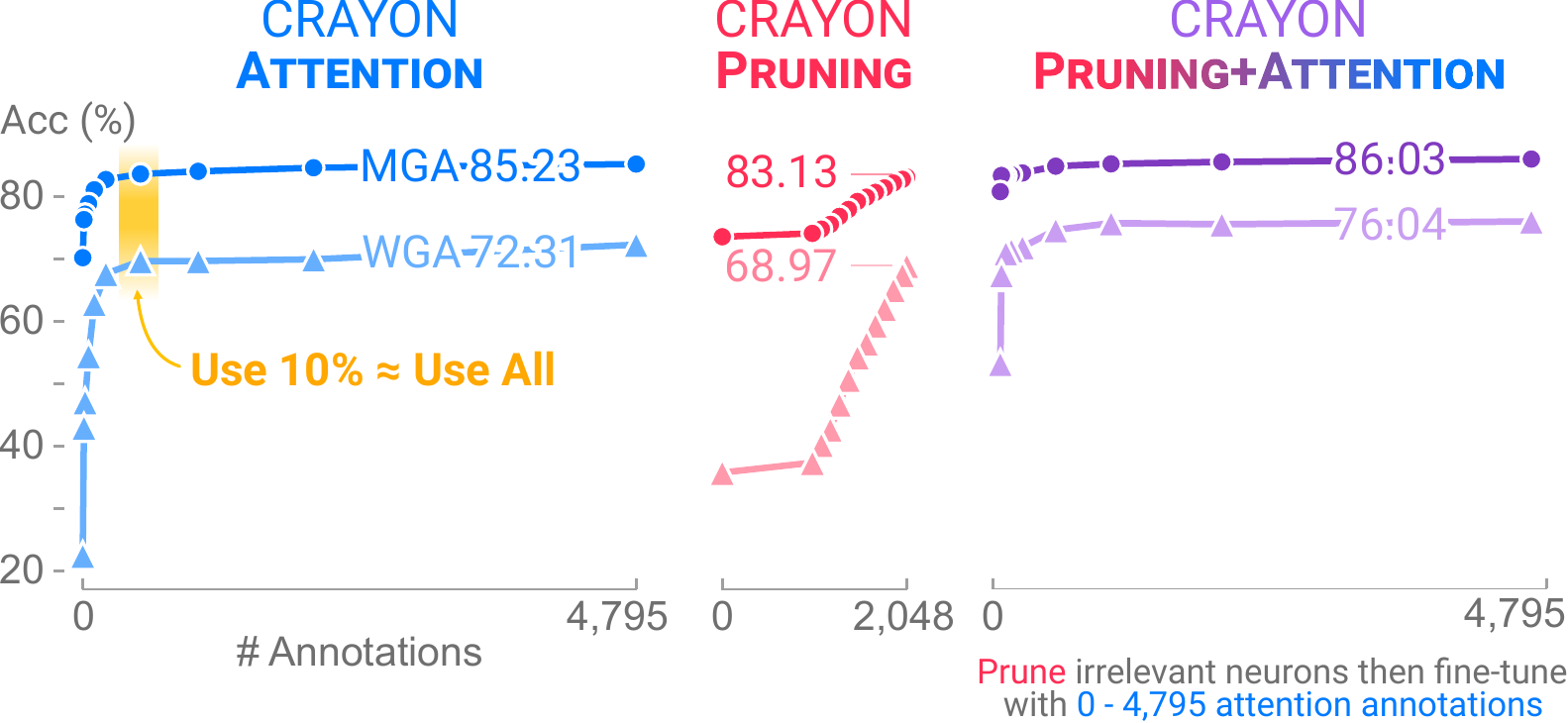}
    \caption{
    For the Waterbirds dataset,
    even with annotations %
    for just 10\% of the training data points, both \methodattention and \methodall nearly reach their peak performance. 
    \methodpruning realizes its full effectiveness when annotations are provided for most neurons in the penultimate layer.
    The performance of \methodall with no attention annotations differs from \methodpruning's peak performance,
    as the latter retrains the entire model, while the former fine-tunes only the last layer.
    For each annotation number $n$,
    we run each method five times with different random seeds and report the average MGA and WGA values.
    }
    \label{fig:num_fb}
\end{figure}

\subsection{Scalability: Runtime Scales Linearly with Data Points}
\label{sec:scalability}
  We run all \method algorithms on an NVIDIA A6000 GPU with 40GB RAM.
  For the Waterbirds dataset with 4,795 training data points, 
  \methodattention takes 288 seconds on average, %
  \methodpruning takes 111 seconds, and %
  \methodall takes 295 seconds. %
  For the Biased CelebA dataset with 20,200 training data points,
  \methodattention takes 1368 seconds,
  \methodpruning takes 1241 seconds, and
  \methodall takes 1401 seconds.
  For the Backgrounds Challenge dataset, which consists of 45,405 training data points,
  \methodattention takes 4512 seconds,
  \methodpruning takes 4118 seconds, and
  \methodall takes 4540 seconds.
  Overall, \methodpruning takes much less time as it fine-tunes only the last fully connected layer.
  Additionally, these runtimes align with our time complexity analysis of \method that it scales linearly with the number of training data points (\autoref{sec:complexity}).

\subsection{Varying Number of Annotations}
\label{sec:num_fb}

We evaluate how the number of annotations $n$ affects \method's performance.
For \methodattention, we randomly sample $n$ images from the training set of the Waterbirds dataset and compute both $\mathcal{L}_{rel}$ and $\mathcal{L}_{irrel}$ for these $n$ images along with their annotations.
For \methodpruning, we randomly sample $n$ neurons from the penultimate layer and 
prune the \textit{irrelevant} neurons within this sampled group of $n$.
We additionally investigate \methodall{}, where we use all 2,048 annotations for neuron relevance
while varying the number of annotations for model attention\footnote{ %
We elect to focus on varying the number of attention annotations %
based on our observation that almost all pruning annotations need to be used for \methodpruning to be fully effective.}. %

\autoref{fig:num_fb} presents the results for  Waterbirds.
The performance of all \method methods improves as the number of annotations increases.
Notably, 
\methodattention and \methodall are effective even with a small number of annotations,
achieving nearly peak performance when annotations are available for only 10\% of the training data points. %
Specifically,
with %
attention annotations for 500 out of 4,795 training data points,
\methodattention enhances the WGA and MGA
to 83.64\% and 69.69\%, respectively,
while \methodall enhances the WGA and MGA to %
84.89\% and 74.67\%, respectively.
These values are only marginally lower than the performance achieved with annotations for all training data points.
In contrast, the performance of \methodpruning is constrained 
unless a substantial portion of neurons is annotated,
underscoring the importance of acquiring annotations for all neurons in the penultimate layer.
In our repository, we include additional results for Biased CelebA and Backgrounds Challenge, which show similar overall trends as presented above for Waterbirds\footnote{\url{https://github.com/poloclub/crayon/blob/main/README.md}}.

\begin{figure*}[t]
    \centering
    \includegraphics[width=0.99\textwidth]{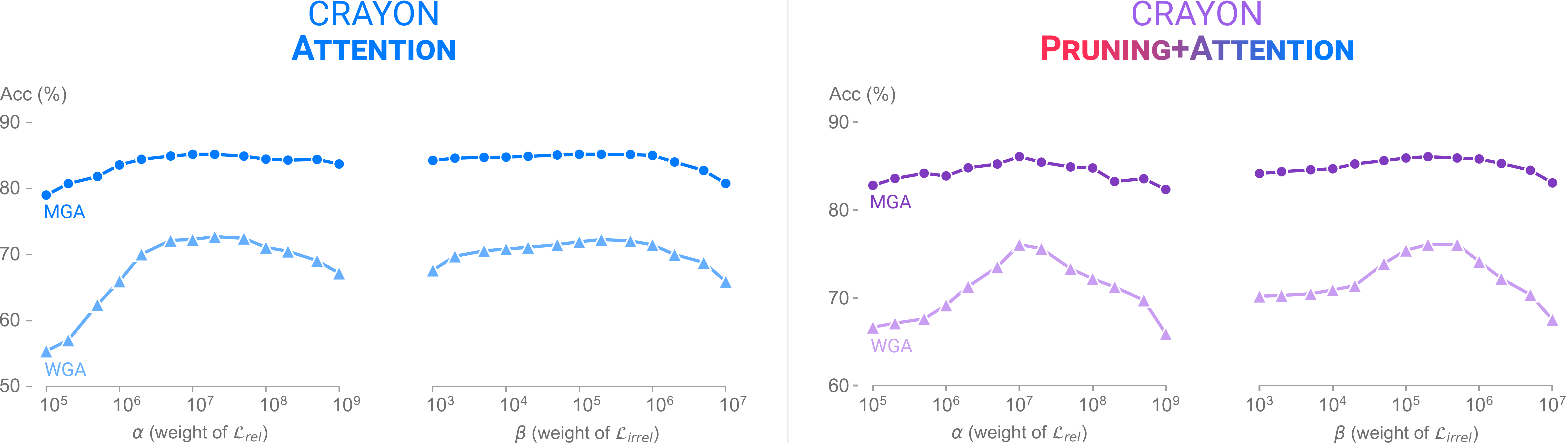}
    \caption{
    For the Waterbirds dataset, 
    both \methodattention (left) and \methodall (right) effectively correct model attention for wide range of $\alpha$ and $\beta$.
    }
    \label{fig:hyperparameter}
\end{figure*}

\subsection{Ablation Study} %
\label{sec:ablation}

\begin{table}[t]
  \addtolength{\tabcolsep}{-3pt}
  \centering
  \caption{
  Ablation study demonstrates that
  $\mathcal{L}_{rel}$ guides model attention significantly stronger than
  $\mathcal{L}_{rel}$ which directs attention away from irrelevant image regions.
  }
  \small 
  \label{tab:ablation}
  \begin{tabular}{@{}c@{\hspace{0.7em}}c@{\hspace{1em}}c@{\hspace{0.7em}}cc@{\hspace{0.7em}}cc@{\hspace{0.7em}}c@{}}
  \toprule
  \multirow{2}{*}[-2pt]{$\mathcal{L}_{rel}$} &  
  \multirow{2}{*}[-2pt]{$\mathcal{L}_{irrel}$} &  
  \multicolumn{2}{@{\hspace{-0.25em}}c}{\textbf{Waterbirds}} & \multicolumn{2}{@{\hspace{0.35em}}c}{\textbf{Biased CelebA}} & \multicolumn{2}{@{\hspace{1em}}c}{\textbf{Backgrounds}} \\ 
  \cmidrule(l{-0.2em}){3-8} 
  && WGA$\uparrow$ & MGA$\uparrow$ & WGA$\uparrow$ & MGA$\uparrow$ & MR$\uparrow$ & BG-Gap$\downarrow$ \\ 
  \midrule
  \checkmark&\checkmark & 72.31 & 85.23 & 83.29 & 89.61 & 80.85 & 8.52 \\
  \checkmark&  & 67.38 & 83.64 & 77.48 & 88.28 & 80.52 & 8.69 \\
  &\checkmark  & 46.57 & 76.33 & 67.32 & 84.59 & 78.66 & 11.86 \\
  & & 22.49 & 70.23 & 37.82 & 71.86 & 77.86 & 12.88 \\
  \bottomrule
  \end{tabular}
  \end{table}

  We conduct an ablation study to assess the impact of the two proposed loss terms, $\mathcal{L}_{rel}$ and $\mathcal{L}_{irrel}$,
  on \methodattention{}'s performance.
  We deactivate one of the two loss terms
  by setting either $\alpha$ or $\beta$ in Equation~\ref{eq:attloss} to 0.
  \autoref{tab:ablation} summarizes the results.
  Overall, removing either loss term degrades the performance of \methodattention,
  highlighting the contributions of both $\mathcal{L}_{rel}$ and $\mathcal{L}_{irrel}$ in guiding model attention.
  When we deactivate $\mathcal{L}_{irrel}$ and rely solely on $\mathcal{L}_{rel}$ (Row~2),
  WGA and MGA experience declines of 
  4.93pp (72.31\% to 67.38\%) and 1.59pp (85.23\% to 83.64\%) for  Waterbirds  and
  5.81pp (83.29\% to 77.48\%) and 1.33pp (89.61\% to 88.28\%) for  Biased CelebA.
  For Backgrounds Challenge,
  the accuracies on the Mixed-Rand datasets decrease 
  from 80.85\% to 80.52\%
  and 
  the BG-Gap increases 
  from 8.52\% to 8.69\%, respectively.
  
  Excluding $\mathcal{L}_{rel}$ from $\mathcal{L}_{att}$ (Row~3) also significantly impairs the performance.
  For  Waterbirds, 
  WGA decreases by 25.74pp (72.31\% to 46.57\%) and 
  MGA by 8.90pp (85.23\% to 76.33\%),
  and
  for Biased CelebA, WGA declines by 15.97pp (83.29\% to 67.32\%) and MGA by 5.02pp (89.61\% to 84.59\%).
  Likewise, for Backgrounds Challenge, 
  the accuracy on the Mixed-Rand dataset decreases by 2.19pp (80.85\% to 78.66\%) and 
  the BG-Gap increases by 3.34pp (8.52\% to 11.86\%).
  These results show 
  that $\mathcal{L}_{rel}$ plays a significant role in guiding model attention, while $\mathcal{L}_{irrel}$ provides additional guidance by directing attention away from irrelevant areas.

\subsection{Transferability of Attention Annotations}
\label{sec:transferability}
Although attention annotations have been collected with respect to the saliency maps from the \textit{Original} model, 
\methodattention can effectively exploit these annotated saliency maps to refine other models,
as these annotated maps serve as indicators of 
where a model should or should not attend.
To assess the transferability of attention annotations,
for each dataset, we train
five ConvNeXt~\cite{liu2022convnet} models, each initialized with a different random seed value.
We then refine these models using \methodattention with the saliency maps from the \textit{Original} model and their relevance annotations.
In other words, we train the ConvNeXt models %
using the saliency maps from the \textit{Original} model as $M_{\mathbf{x}_n}$ in \autoref{eq:rel_loss} and \ref{eq:irrel_loss},
and their relevance annotations to decide $R$ and $I$ in \autoref{eq:attloss}.
We note that $M_{\mathbf{x}_n}'$ in \autoref{eq:rel_loss} and \ref{eq:irrel_loss} is the saliency map for the ConvNeXt models being trained. %
We run the \methodattention for each model five times, each with a different random seed, and report the average performance in \autoref{tab:transfer}.

The attention annotations for the \textit{Original} model's saliency maps effectively guide the attention of the other models.
For the Waterbirds dataset,
the average WGA of the five models, which was 66.23\% before refining, increases to 76.73\%,
and the average MGA, which was 84.73\%, is elevated to 88.59\%.
Similarly, for the Biased CelebA dataset, the average WGA improves from 26.76\% to 49.50\%, and MGA from 71.32\% to 79.40\%.
For the Backgrounds dataset, the average accuracy on the Mixed-Rand dataset is enhanced from 87.18\% to 89.49\%, and the BG-Gap is reduced from 5.87\% to 5.56\%.
These results demonstrate that our attention annotations can be effectively transferred across different models.

\begin{table}[h]
\addtolength{\tabcolsep}{-2pt}
\centering
\vspace{-3pt}
\caption{
The attention annotations for the \textit{Original} ResNet50 model effectively refine ConvNeXt trained on 3 datasets.
}
\small 
\label{tab:transfer}
\begin{tabular}{@{}lcccccc@{}}
\toprule
\multirow{2}{*}{} & \multicolumn{2}{c}{\textbf{Waterbirds}} & \multicolumn{2}{c}{\textbf{Biased CelebA}} & \multicolumn{2}{c}{\textbf{Backgrounds}} \\
\cmidrule{2-7} 
& WGA$\uparrow$ & MGA$\uparrow$ & WGA$\uparrow$ & MGA$\uparrow$ & MR$\uparrow$ & BG-Gap$\downarrow$  \\ 
\midrule
Original & 66.23 & 84.73 & 26.76 & 71.32 & 87.18 & 5.87 \\
Refined  & 76.73 & 88.59 & 49.50 & 79.40 & 89.49 & 5.56 \\
\bottomrule
\end{tabular}
\end{table}

\subsection{Hyperparameter Sensitivity}
\label{sec:hyperparam}

  We examine how 
  \method's two hyperparameters, $\alpha$ and $\beta$,
  affect the performance of \methodattention and \methodall.
  We evaluate the performance of the two methods on the Waterbirds dataset with different values of $\alpha$ over the wide range from 1e+5 to 1e+9 while consistently using $\beta$ of 2e+5 as in \autoref{sec:crayonsetup}.
  Similarly, we vary $\beta$ from 1e+3 to 1e+7, while keeping $\alpha$ as 1e+7.
  \autoref{fig:hyperparameter} visualizes the results.

  We observe that 
  the performance of \methodattention stabilizes for various values of $\alpha$ from 5e+6 to 5e+7.
  The performance is even more stable with different values of $\beta$ ranging from 2e+3 to 5e+5.
  The performance of \methodall is relatively more sensitive to these hyperparameters;
  changing $\alpha$ by a scale of 2 can impact the performance, while the performance stabilizes for $\beta$ from 1e+5 to 5e+5.
  This demonstrates that both \methodattention and \methodall achieve its effectiveness with various values of the hyperparameters, while they would benefit from a careful hyperparameter tuning.
  Our repository provides the results of the hyperparameter sensitivity experiments for the Biased CelebA and Backgrounds Challenge dataset, which show similar trends to the Waterbirds dataset\footnote{\url{https://github.com/poloclub/crayon/blob/main/README.md}}.

\section{Reproducibility}
\label{sec:checklist}
Our research is fully reproducible. We have carefully prepared and described all necessary information for interested researchers to reproduce and extend our work. Our work has met all the criteria outlined in the required reproducibility checklist. Below, we summarize how we have ensured reproducibility. We reference specific sections and subsections of this paper, and specific files in our repository, to help readers easily locate the information they need for reproducibility.

We have clearly described the settings of all presented algorithms (\autoref{sec:crayonsetup}; \verb|misc/training_config.md| in repository) and all models used (\autoref{sec:dataset}, \autoref{sec:transferability}).  
Additionally, we have analyzed their time and space complexity (\autoref{sec:complexity}) and how they scale with increasing sample sizes (\autoref{sec:crayonsetup}).

For all three datasets used in our paper, we have described essential information, including the number of data points in training and test datasets, their distributions, and image pre-processing steps (\autoref{sec:dataset}). 
We have also detailed the process of collecting the \yesnof for \method by describing the visualizations and questions presented to the annotators and how we integrated the collected data into high-quality annotations (\autoref{sec:annotation}).
In our repository,
we provide the links to download the three datasets and \yesnof (\verb|README.md|).

To facilitate easy reproduction, 
we have included
specifications of dependencies (\verb|environment.yml|),
training and evaluation code (\verb|src/solver|),
a link to download pretrained models to run the code, and
a README file with a table of experiment results and comprehensive guidelines on how to reproduce the results (\verb|README.md|).

For the experiments,
we have clarified the specification of all hyperparameters and how we selected the values (\autoref{sec:crayonsetup}, \autoref{sec:hyperparam}, \verb|misc/training_config.md|).
We have demonstrated 
the number of runs for training and evaluation (\autoref{sec:performance}),
the definition of the used metrics (\autoref{sec:metrics}), and
the mean and standard deviation of the results (\autoref{tab:perf}).
We have also reported 
the details about the computing resources used for the experiments and the average runtime of each experiment in \autoref{sec:crayonsetup}.

This work does not aim to propose theoretical claims; therefore, theoretical proofs and statements are not included.

\section{Conclusion}
\label{sec:conclusion}
We introduce \method, the state-of-the-art approach for correcting model attention using %
simple \yesnof annotations. 
Extensive quantitative evaluation with large-scale human annotations
demonstrates \method's effectiveness, scalability, and practicality. 
It outperforms all existing methods across three benchmark datasets, surpassing approaches that rely on more complex annotations.

\bibliographystyle{IEEEtran}
\bibliography{reference}
\end{document}